\pdfoutput=1
\documentclass[11pt]{article}

\usepackage[final]{acl}

\usepackage{times}
\usepackage{latexsym}

\usepackage[T1]{fontenc}
\usepackage[utf8]{inputenc}

\usepackage{microtype}

\usepackage{inconsolata}

\usepackage{graphicx}

\usepackage{adjustbox}
\usepackage{float}
\usepackage{placeins}

\usepackage{soul}
\usepackage{paralist}
\usepackage{amsmath, amsfonts}
\usepackage{bm}
\usepackage[capitalise,noabbrev]{cleveref}
\usepackage{CJKutf8}
\usepackage{enumitem}

\usepackage{booktabs}
\usepackage{array} 
\usepackage{multirow}
\usepackage{calc}	
\usepackage{makecell}

\newcolumntype{C}[1]{>{\centering\arraybackslash}p{#1}}

\title{How to Make the Most of LLMs’ Grammatical Knowledge\\for Acceptability Judgments}

\author{
    \textbf{Yusuke Ide}\;\;\;
    \textbf{Yuto Nishida}\;\;\;
    \textbf{Justin Vasselli}\;\;\;
    \textbf{Miyu Oba}\;
    \\
    \textbf{Yusuke Sakai}\;\;\;
    \textbf{Hidetaka Kamigaito}\;\;\;
    \textbf{Taro Watanabe}
    \\
    Nara Institute of Science and Technology
    \\
    \texttt{\{ide.yusuke.ja6, nishida.yuto.nu8, vasselli.justin\_ray.vk4, oba.miyu.ol2,} 
    \\
    \texttt{sakai.yusuke.sr9, kamigaito.h, taro\}@is.naist.jp} \\
}

\begin{document}
\begin{CJK}{UTF8}{gbsn}

\maketitle

\begin{abstract}
The grammatical knowledge of language models (LMs) is often measured using a benchmark of linguistic minimal pairs, where the LMs are presented with a pair of acceptable and unacceptable sentences and required to judge which is more acceptable.
Conventional approaches directly compare sentence probabilities assigned by LMs, but recent large language models (LLMs) are trained to perform tasks via prompting, and thus, the raw probabilities they assign may not fully reflect their grammatical knowledge. 
% By prompting LLMs, using prompts or templates, to explicitly evaluate grammaticality, we can potentially obtain more accurate acceptability judgments.
% Conventional approaches compare sentence probabilities directly, but they may not be the optimal way to make use of the grammatical knowledge of large language models (LLMs).
% By providing LLMs with targeted guidance, we can potentially obtain more accurate acceptability judgments.
% We therefore investigate how to derive the most accurate acceptability judgments from LLMs to comprehensively evaluate their grammatical knowledge.

In this study, we attempt to derive more accurate acceptability judgments from LLMs using prompts and templates.
Through extensive experiments in English and Chinese, we compare nine judgment methods and find two of them, a probability readout method---\textit{in-template LP} and a prompt-based method---\textit{Yes/No probability computing}, achieve higher accuracy than the conventional ones.
Our analysis reveals that these methods excel in different linguistic phenomena, suggesting they access different aspects of LLMs' knowledge.
We also find that ensembling the two methods outperforms single methods.
Consequently, we recommend these techniques, either individually or ensembled, as more effective alternatives to conventional approaches for assessing grammatical knowledge in LLMs.
\footnote{
    Our codes and templates are published at \url{https://github.com/Yusuke196/llm-acceptability}.
}
\end{abstract}

%%%%%%%%%%%%%%%%%%%%%%%%%%%%%%%%%%%%%%%%%%%%%%%%%%

\section{Introduction}
\label{sec:intro}

The grammatical knowledge of language models (LMs) is often measured using acceptability judgments \citep{Lau2017-ye,warstadt-etal-2019-neural}.
% Note: Acceptability judgments are also called grammaticality judgments (Schütze, 1996)
There are two main categories of acceptability judgment benchmarks, the single-sentence one and the minimal-pair (MP) one, as detailed in \cref{sec:related}.
We focus on the latter because it allows us to directly measure LMs' grammatical knowledge without task-specific fine-tuning.
Below is an example of a minimal pair from \citet{warstadt-etal-2020-blimp}.
\begin{asparaenum}[(a)]
    \item \textit{These casseroles \ul{disgust} Kayla.}
    \item \textit{*These casseroles \ul{disgusts} Kayla.}
\end{asparaenum}
Here, sentence (a) is acceptable or grammatically correct, while (b) is not, as its underlined verb violates the subject-verb agreement.
% MPP benchmarks can evaluate any LMs including ones that have only been pre-trained, e.g., with next token prediction \cite{mueller-etal-2020-cross,zhang-etal-2021-need,hewitt-etal-2023-backpack}.
% As such, MP benchmarks can evaluate any LMs including ase models and nstruct models, without fine-tuning for acceptability judgments.

Meanwhile, the recent scaling up of model sizes and training data for LMs has made it possible to solve a wide range of tasks using few-shot or zero-shot prompting, without the need for task-specific fine-tuning \cite{Brown2020-za,Liu2021-ou}, popularizing the term large language model (LLM).
Incorporating learning techniques such as instruction tuning \cite{wei2022finetuned} and Direct Preference Optimization (DPO) \cite{rafailov2023direct} further improved the alignment of LLM outputs with human preferences.
The LLMs trained by these techniques achieve good performance by prompting, i.e., guidance on what knowledge to elicit.

In this light, various methods can be developed to obtain more accurate acceptability judgments by providing LLMs with targeted guidance.
% no previous studies have thoroughly explored them;
However, as discussed in \cref{sec:related}, most previous studies simply input the sentences into the (L)LM, calculate their probabilities, and consider the sentence with the higher probability as the acceptable one.
Although \citet{hu-levy-2023-prompting} compared multiple methods of obtaining acceptability judgments from LLMs (see their Experiment 3b), their probability readout method and prompting method were limited to basic ones.
As a result, they broadly claim that prompting is ineffective, whereas our experiments demonstrate that it can be highly effective with the proper technique.
%their probability readout method remained the conventional one, and their prompting method was also limited to the basic one asking LLMs to respond with either of the given choices, 1 or 2.
% This suggests the potential for developing more accurate methods to derive judgments from LLMs.
%This leaves room for exploring methods that derive more accurate judgments from LLMs.
% This leaves open the question of which methods derive the most accurate acceptability judgments from LLMs and what the pros and cons of each method are.
% Consequently, it is unclear what methods are effective in obtaining acceptability judgments using LLMs and what their strengths and weaknesses are.
% They simply input the given sentences to an LM, calculate their probabilities, and deem the sentence with the higher probability of the pair to be acceptable according to the LM.
% Such a method is problematic in that the probabilities can be influenced by aspects irrelevant to grammatical acceptability, e.g., sentence length \citep{lau-etal-2020-furiously}.
% Normalized measures such as PenLP \cite{Wu2016-rp,lau-etal-2020-furiously} have been shown to mitigate some of this bias, but they do not eliminate it \cite{ueda-etal-2024-token-length}.
% At the same time, prompting large language models (LLMs), which is shown to be strong in many other tasks \cite{Liu2021-ou}, has yet to be thoroughly examined for acceptability judgments.

\begin{figure}[t]
    \includegraphics[width=1.0\linewidth]{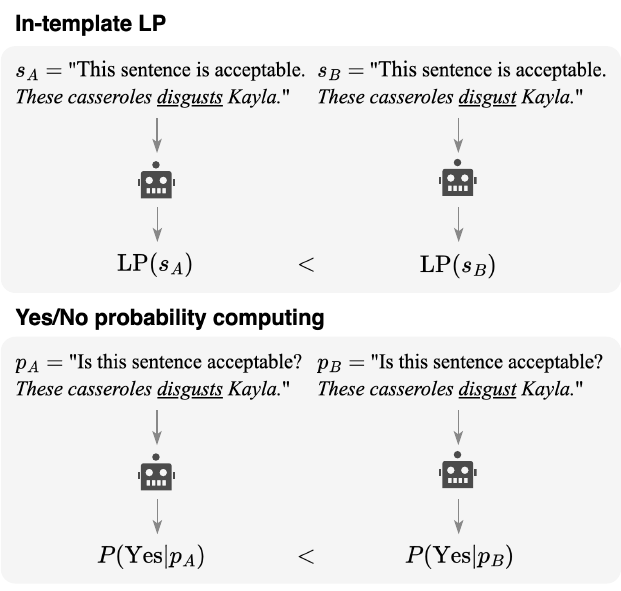}
    
    \caption{Conceptual illustration of our top methods. Differences between paired sentences are underlined. Both methods judge the sentence that results in a higher (log) probability acceptable. See \cref{sec:methods} for the details.}
    \label{fig:concept-diagram}
\end{figure}

We contribute to this area by comparing (1) the conventional sentence probability readout\footnote{
    Readout refers to accessing an LLM's output layer to compute probabilities of strings \cite{kauf2024comparingplausibilityestimatesbase}. % This is not a verbatim citation
} methods, (2) our novel probability readout methods in \textit{in-template} settings, and (3) prompt-based methods.
In the in-template probability readout, we insert each sentence into a template and use the LLM to calculate the probability of the complete string. 
The template allows us to guide the LLM to judge the sentence's grammaticality in a way that conventional probability readouts cannot.
%In the in-template probability readout, we insert each target sentence into a template before inputting it into an LLM, instructing the LLM to focus on its grammaticality.
We call the most basic method \textit{in-template LP}, where LP stands for log probability.
For prompt-based methods, we investigate a basic method of asking LLMs to respond with a choice and \textit{Yes/No probability computing (Yes/No prob comp)}, where we compute the normalized probability of ``Yes'' versus ``No'', inspired by UniEval \cite{zhong-etal-2022-towards}. %, which is shown to correlate well with humans in evaluating natural language generation.
\cref{fig:concept-diagram} presents the conceptual illustration of in-template LP and Yes/No prob comp.

To rigorously compare these methods, we conduct experiments using eight LLMs and two MP benchmarks (one for English and one for Chinese).
% First, an in-template probability readout method, \textit{in-template LP},\footnote{
%     LP stands for log probability.
% } and Yes/No prob comp (See \cref{fig:concept-diagram} for their conceptual illustration) show top performance, surpassing the conventional methods. 
The results show the effectiveness of the two methods.
% The results show that the two best-performing methods are (1) in-template LP and (2) Yes/No prob comp.
% The results show the excellence of two methods, an in-template probability readout method, \textit{in-template LP},\footnote{
%     LP stands for log probability.
% } and Yes/No prob comp. 
% See \cref{fig:concept-diagram} for their conceptual illustration.
% These two methods achieve the highest accuracy on the English and Chinese benchmark, respectively, mostly surpassing the conventional methods (\cref{sec:results}).
% Yes/No prob comp and in-template LP achieve the highest accuracy on the English benchmark and Chinese benchmark, respectively, outperforming conventional methods in all but one setting (\cref{sec:results}).
In-template LP consistently outperforms conventional methods, achieving the highest accuracies on the Chinese benchmark.
Yes/No prob comp achieves the highest accuracies on the English benchmark in all but one setting.
% We also find that the basic prompting method of making LLMs select an answer from multiple choices (A/B prompting) mostly underperforms the other prompting method, Yes/No prob comp, indicating ... % TODO?

% \begin{enumerate}[itemsep=0pt, parsep=0pt]
%     \item An in-template probability readout method, \textit{in-template LP},\footnote{
%         LP stands for log probability.
%     } and Yes/No prob comp (See \cref{fig:concept-diagram} for their conceptual illustration) show top performance, surpassing the conventional methods.
%     \item In-template LP and Yes/No prob comp have different strengths; for example, Yes/No prob comp is robust against token-length bias. 
%     This indicates that they harness different aspects of LLMs' grammatical knowledge, helping comprehensive evaluation of LLMs.
%     \item Ensembling the two methods further improves the accuracy, revealing their complementary capabilities. 
%     The highest score by \textit{Mix-P3} with Qwen2 is 1.6 percentage points higher than humans on the English benchmark.
%     \item Even with the top two methods, all the LLMs have trouble making correct judgments where the unacceptable sentence can be obtained by shuffling the words in the acceptable one.
%     % The accuracy for such paradigms averaged over models and methods is only 71.6\% compared to 87.9\% of other paradigms. % paradigmという単語をまだ導入していないので、ここでは使えない
% \end{enumerate}
% In conclusion, we recommend employing diverse judgment methods instead of relying on conventional sentence probability readout methods.

Moreover, our analysis demonstrates the following key findings.
(1) In-template LP and Yes/No prob comp have different strengths; for example, Yes/No prob comp is robust against token-length bias.
This indicates that they access different aspects of LLMs' grammatical knowledge, contributing to a more comprehensive evaluation.
(2) Ensembling the two methods further improves the accuracy, revealing their complementary capabilities. 
The highest score, achieved with Qwen2, is 1.6 percentage points higher than humans on the English benchmark.
% Based on these findings, we recommend the following to make the most of LLMs' grammatical knowledge: first, use Yes/No prob comp and in-template LP; then, try ensembling the two methods if possible.
Based on these findings, we recommend the following: if possible, ensemble the two methods; otherwise, use in-template LP.
(3) We identify a common weakness across all our combinations of the LLMs and methods: they struggle to make correct judgments on linguistic phenomena where the unacceptable sentence can be obtained by shuffling the words in the acceptable one, which presents a challenge for future work.

%%%%%%%%%%%%%%%%%%%%

\section{Related Work}
\label{sec:related}

% Maybe introduce the acceptability judgments for humans citing chomsky1957syntactic here
Benchmarks of acceptability judgments can be divided into two categories: single-sentence benchmarks and MP benchmarks.
The single-sentence benchmarks pose a binary classification of single sentences as seen in CoLA \cite{warstadt-etal-2019-neural}, a dataset composed of sentences each labeled acceptable or unacceptable.
CoLA was incorporated into the natural language understanding benchmark GLUE \cite{wang-etal-2018-glue} and has been used to evaluate a wide range of models, including LMs.
However, single-sentence benchmarks cannot measure LMs' grammatical knowledge directly because they require training a supervised classifier before the evaluation. 
This makes it difficult to distinguish between the knowledge of the model itself and what is learned by training the classifier \cite{warstadt-etal-2020-blimp}.

%In contrast, MP benchmarks do not need task-specific training as they present minimally different pairs, asking which is acceptable.
In contrast, MP benchmarks present minimally different pairs and the task is to determine which is the most acceptable sentence, eliminating the need for a classifier.
As another advantage of the MP benchmark, minimal pairs can be automatically generated in a controlled manner, providing a sufficient amount of quality data for model evaluation \cite{linzen-etal-2016-assessing}. % MP benchmarks like JBLiMP are not automatically generated.
In conventional experiments using an MP benchmark, judgments are made based on sentence probabilities.
Models are evaluated by whether they assign a higher probability to the acceptable sentence in each minimal pair.
This method, which we call sentence probability readout, has been dominantly employed for MP acceptability judgments across languages (\citealp{marvin-linzen-2018-targeted,warstadt-etal-2020-blimp,mueller-etal-2020-cross,haga-etal-2024-modeling,xiang-etal-2021-climp,someya-oseki-2023-jblimp}, inter alia).

Experiments with prompting LLMs have been conducted on both single-sentence benchmarks and MP benchmarks.
\citet{zhang-etal-2024-mela} compared various models, including LLMs, on a single-sentence benchmark.
On MP benchmarks, \citet{hu-levy-2023-prompting} compared the sentence probability readout and prompting.
% semantic plausibilityの話はおそらく不要
% Specifically, they conducted experiments using large language models (LLMs) on two MPP tasks: acceptability judgment and semantic plausibility judgment.
% Semantic plausibility judgment is the task of determining which of the given words is more likely, given the preceding context.
% The LLMs they employed were three Flan-T5 models of different sizes, one GPT-3 model, and two GPT-3.5 models. 
% They conducted experiments using LLMs and an MP benchmark composed of subsets of BLiMP \cite{warstadt-etal-2020-blimp} and SyntaxGym \cite{gauthier-etal-2020-syntaxgym} to show that sentence probability readout generally outperforms prompting.
% However, their probability readout method and prompting method were limited to basic ones, and they did not analyze the accuracy difference between methods in detail, e.g., according to linguistic phenomena.
However, their probability readout and prompt-based methods relied on basic implementations, without systematically exploring alternative ways to calculate sentence probabilities or optimize prompting strategies.

% Another line of work has revealed that the token length influences the performance of sentence probability readout; normalized measures such as PenLP \cite{Wu2016-rp} have been shown to mitigate some of this bias \cite{lau-etal-2020-furiously}, but they do not eliminate it \cite{ueda-etal-2024-token}.
Another line of work has studied biases that affect the performance of sentence probability readout.
For example, it is known that the sentence probabilities given by LMs tend to decline as the token length of the sentence grows; normalized measures such as PenLP \cite{Wu2016-rp} have been shown to mitigate some of this bias \cite{lau-etal-2020-furiously}, but they do not eliminate it \cite{ueda-etal-2024-token}.

%%%%%%%%%%%%%%%%%%%%

\begin{table*}[ht]
    \centering
    \small
    \begin{tabular}{l p{0.72\linewidth}}
        \toprule
        Input Type & Example Input \\
        \midrule
        Sentence & \textit{Many girls insulted themselves.} \\
        \cmidrule(r){1-1} \cmidrule(l){2-2}
        In-template single & The following sentence is grammatically acceptable.\textbackslash n\textbackslash n\textit{Many girls insulted themselves.} \\
        \cmidrule(r){1-1} \cmidrule(l){2-2}
        \multirow{2}{*}{In-template comparative} & The following sentence A is grammatically acceptable while B is not.\textbackslash n\textbackslash nA: \textit{Many girls insulted themselves.}\textbackslash nB:\textit{Many girls insulted herself.} \\
        \bottomrule
    \end{tabular}
    \caption{Example English inputs of the readout methods. The target or inserted sentences are in italics. See \cref{tab:template-settings-zh} for Chinese versions.}
    \label{tab:template-settings}
\end{table*}

\begin{table*}[ht]
    \centering
    \small
    \begin{tabular}{l l p{0.76\linewidth}}
        \toprule
        Type & Role & Example Message \\

        \midrule
        \multirow{4}{*}{A/B} & System & Your task is to compare the quality of given sentences. \\
        % \cmidrule(r){1-1} \cmidrule(l){2-2}
        \cmidrule(rl){2-2} \cmidrule(l){3-3}
        & \multirow{3}{*}{User} & One of the following sentences is grammatically acceptable and the other is not. Which one is acceptable? Respond with A or B as your answer.\textbackslash n\textbackslash nA: \textit{Many girls insulted themselves.}\textbackslash nB: \textit{Many girls insulted herself.} \\

        \midrule
        \multirow{4}{*}{Yes/No} & System & Your task is to evaluate the quality of given text. \\
        \cmidrule(rl){2-2} \cmidrule(l){3-3}
        & \multirow{2}{*}{User} & Is the following sentence grammatically acceptable? Respond with Yes or No as your answer.\textbackslash n\textbackslash n\textit{Many girls insulted themselves.} \\
        \bottomrule
    \end{tabular}
    \caption{Example English messages for prompting. The target or inserted sentences are in italics. See \cref{tab:prompts-zh} for Chinese versions.}
    \label{tab:prompts}
\end{table*}

%%%%%%%%%%%%%%%%%%%%

\section{Methods}
\label{sec:methods}

We compare three different groups of methods to extract acceptability judgments from the LLMs.

\subsection{Sentence Probability Readout}

In sentence probability readout, we input each sentence of a given pair into a model to obtain the probabilities assigned to each token.
The probabilities are then used to compute a probability score for each sentence, and the sentence given the higher score is predicted to be acceptable.
% Then we compute the probability score of each sentence using a measure from the probabilities and predict the sentence given a higher score as acceptable.

We experiment with three measures to compute the probability scores: LP, MeanLP, and PenLP.
All of them have been widely used in acceptability judgments.\footnote{
    SLOR \cite{kann-etal-2018-sentence} is also commonly used, but it requires building a unigram model using the training data of the LM.
    Because the training data of the LLMs we use is not publicly available, we skip the acceptability measure.
}
LP is the unnormalized log probability of the sentence
\begin{align}
    \mathrm{LP}(\bm{s}) &= \log P(\bm{s})
\end{align}
where $\bm{s}$ is the input sequence of tokens and $P(\bm{s})$ is the probability assigned to $\bm{s}$ by the model
\begin{align}
    P(\bm{s}) &= \prod_{i = 1}^{|\bm{s}|} P(t_i | t_{<i}).
\end{align}
Because LP tends to get smaller as the sentence gets longer \cite{ueda-etal-2024-token}, we also compute two normalized measures, MeanLP and PenLP \cite{lau-etal-2020-furiously,Wu2016-rp}, 
% While sentence length can be defined either by the number of tokens or by the number of words, we consider normalization based on the number of tokens. 
\begin{align}
    \mathrm{MeanLP}(\bm{s}) &= \frac{\log P(\bm{s})}{|\bm{s}|} \\
    \mathrm{PenLP}(\bm{s}) &= \frac{\log P(\bm{s})}{((5 + |\bm{s}|)/(5 + 1))^\alpha}
\end{align}
where $\alpha$ is a hyperparameter to scale the sentence length, reducing the impact of long sentences and ensuring a fair comparison across different lengths.
We set $\alpha = 0.8$ following \citet{lau-etal-2020-furiously} and \citet{ueda-etal-2024-token}.
We hereafter refer to the three judgment methods simply by the name of the corresponding measures: \textit{LP}, \textit{MeanLP}, \textit{PenLP}.

\subsection{In-template Probability Readout}

% We propose a new judgment approach, in-template probability readout, that explicitly guides the model to focus on the acceptability or grammaticality of the sentence.
% This is done by inserting sentences into a template to build an input string.
In-template probability readout follows the same steps of computing and comparing probabilities as sentence probability readout.
Meanwhile, its input string is built by embedding the sentences in a template designed to draw focus to their grammaticality.
% Instead of calculating the probability scores of the sentences directly, we calculate the probability scores of the sentences embedded in a template designed to draw focus to the grammaticality of the sentences.
% We include guidance to focus on the acceptability or grammaticality of the sentence in the input, which is done by inserting one or two sentences into a template.
The input has two types: \textit{in-template single} and \textit{in-template comparative}.
For each type, we prepare five templates per language because the performance can vary due to minor differences in expressions within prompts \cite{zheng2023judging}.
The templates were created based on those of Flan\footnote{
    \url{https://github.com/google-research/FLAN}
}\cite{wei2022finetuned}.
For Chinese experiments, we use translations of English templates.
Translations were generated by DeepL\footnote{\url{https://www.deepl.com}} and post-edited by a native Chinese speaker.

\paragraph{In-template single}
In-template single templates have one placeholder where the target sentence is inserted.
\cref{tab:template-settings} shows an example input.

As the length of the input changes depending on the length of the target sentence, predictions by in-template single inputs must also use normalization techniques.
%Similarly to the sentence probability readout, predictions by in-template single inputs depend on normalization techniques, as the techniques can change which of the paired sentences gets a higher probability score.
We thus apply each of the three measures explained above to the method, dubbing the corresponding methods \textit{in-template LP}, \textit{in-template MeanLP}, and \textit{in-template PenLP}, respectively.
The final measure depends on whether we base our computations on the whole input string or the target sentence only.
We report the result of the former because it performed better in our preliminary experiments.

\paragraph{In-template comparative} 
In-template comparative inputs are built by filling two placeholders; we insert the target sentence into the first one and the other sentence of the minimal pair into the second.
\cref{tab:template-settings} shows an example input.
Note that the second sentence is supplementary, and our aim here is to measure the acceptability of the first one.

In-template comparative does not need normalization, because the sum of the token lengths of two sentences and, thus, the length of the whole input string is constant, no matter which of the paired sentences comes first.
% Thus, the only measure we employ with the in-template comparative input is LP. 
% We refer to this method as \textit{in-template comparative LP}.
Hence, we only calculate LP for the in-template comparative input, referring to this method as \textit{in-template comparative LP}.

\subsection{Prompt-based Methods}

In prompt-based methods, \textit{A/B prompting} and \textit{Yes/No prob comp}, we provide the models with prompts that include a question.
For both methods, we prepare a system message and a user message.
The system message describes the task to be solved, which has been shown to enhance the performance \cite{peng-etal-2023-towards}.
The user message includes the main prompt, and we prepare five versions of each method's prompt template per language.
Each user message is built by inserting one or two sentences into a template, as we do for in-template probability readout.
When prompting a base model, we concatenate the two messages and append the string \texttt{\textbackslash nAnswer:} at the end.
When prompting an instruct model, we apply chat templates\footnote{
    \url{https://huggingface.co/docs/transformers/en/chat_templating}
} to maximize the performance, including control tokens like \texttt{<|begin\_of\_text|>} in the inputs to the model. 
For Chinese experiments, we use translations of English templates verified by a native Chinese speaker.

\paragraph{A/B prompting}
A/B prompting inputs a prompt containing the paired sentences to the models and asks which sentence is acceptable.
The prompt is exemplified in \cref{tab:prompts}.
The user message contains one acceptable and one unacceptable sentence.
Their order (which sentence goes to A or B) is randomized to eliminate the potential bias from the order \cite{pezeshkpour2023largelanguagemodelssensitivity}.
We perform constrained decoding by outlines\footnote{
    \url{https://github.com/outlines-dev/outlines}
} \cite{willard2023efficient} to ensure that the model outputs either A or B, because our preliminary experiments without outlines observed many outputs violating the constraint.
We turn off sampling in decoding.

\paragraph{Yes/No probability computing}
In Yes/No prob comp, we compute the score of each sentence as the normalized probability of ``Yes'' versus ``No'' given a prompt asking its acceptability.
An example prompt is shown in \cref{tab:prompts}.
We predict the sentence that resulted in a higher ``Yes'' probability to be acceptable.
This method is inspired by UniEval \cite{zhong-etal-2022-towards}, which is shown to correlate well with human judgments in evaluating natural language generation.
We formulate the probability given a sentence $\bm{s}$ as follows,
\begin{align}
    % P(\mathrm{``Yes"}|\bm{s}) = \frac{logit(\mathrm{``Yes"}|\bm{s})}{logit(\mathrm{``Yes"}|\bm{s}) + logit(\mathrm{``No"}|\bm{s})}
    P(\mathrm{``Yes"}|\bm{s}) \!=\! \frac{P_{\mathrm{LLM}}(\mathrm{``Yes"}|\bm{s})}{P_{\mathrm{LLM}}(\mathrm{``Yes"}|\bm{s}) + P_{\mathrm{LLM}}(\mathrm{``No"}|\bm{s})}
\end{align}
where $P_{\mathrm{LLM}}(\cdot)$ is the probability of a word assigned by the model. % word $\bm{w}$
% \begin{align}
%     P_{\mathrm{LLM}}(\bm{w} | \bm{s}) &= \prod_{i = 1}^{|\bm{w}|} P_{\mathrm{LLM}}(t_i | \bm{s}, t_{<i}).
% \end{align}
% where $logit(\cdot)$ is the logit of a token outputted from the final layer of the model. # アーカイブ
For the Chinese experiments, we substitute ``是'' and ``否'' for ``Yes'' and ``No'', respectively, if these words are not segmented into subwords; otherwise, we employ the same formulation as English experiments.
% \footnote{
%     This is the case for Qwen2 and Qwen2-Instruct. 
%     % In our preliminary experiments, requiring 是/非 performed better on Qwen2-Instruct and requiring Yes/No performed better on Qwen2.
% }; otherwise, we employ the same formulation as English experiments.
% Archive:
% In all our experiments, no tokenizers segment these words into subwords.
% Although tokenizers sometimes segment $\bm{w}$ into subwords, this causes no length biases because the yes word and no word turn into the same number of subwords in all our experiments. % all our tokenizers segment those words into sequences of the same length. % segmentation occurs on ``是'' and ``否'' only

%%%%%%%%%%%%%%%%%%%%

\begin{table*}[ht]
    \centering
    \small

    \begin{tabular}{
        >{\hspace{-2pt}}l<{\hspace{-2pt}}
        c<{\hspace{-2pt}}
        c<{\hspace{-2pt}}
        c<{\hspace{-2pt}}
        c<{\hspace{-2pt}}
        c<{\hspace{-2pt}}
        c<{\hspace{-2pt}}
    }
    \toprule

     & Llama-3 & Llama-3-Inst. & Mixtral & Mixtral-Inst. & Qwen2 & Qwen2-Inst. \\
    \midrule
    LP & 79.6 & 77.1 & 82.5 & 82.3 & 80.4 & 79.7 \\
    MeanLP & 77.1 & 74.8 & 79.6 & 79.4 & 77.7 & 77.1 \\
    PenLP & 79.2 & 76.8 & 82.2 & 82.0 & 79.9 & 79.2 \\

    \midrule
    In-template LP & $\mathbf{84.4}$$^{*}$$_{\pm 0.5}$ & \underline{83.5}$^{*}$$_{\pm 0.5}$ & \underline{84.0}$^{*}$$_{\pm 0.5}$ & \underline{83.5}$^{*}$$_{\pm 0.9}$ & \underline{83.9}$^{*}$$_{\pm 0.3}$ & 80.1$^{*}$$_{\pm 1.0}$ \\
    In-template MeanLP & 82.6$^{*}$$_{\pm 0.7}$ & 81.9$^{*}$$_{\pm 0.5}$ & 82.6$^{*}$$_{\pm 0.3}$ & 82.2$_{\pm 0.8}$ & 82.0$^{*}$$_{\pm 0.7}$ & 78.7$_{\pm 1.1}$ \\
    In-template PenLP & \underline{83.8}$^{*}$$_{\pm 0.5}$ & 83.0$^{*}$$_{\pm 0.5}$ & 83.8$^{*}$$_{\pm 0.4}$ & 83.3$^{*}$$_{\pm 1.0}$ & 83.2$^{*}$$_{\pm 0.4}$ & 79.8$_{\pm 1.1}$ \\
    In-template compar. LP & 71.8$_{\pm 4.5}$ & 61.8$_{\pm 2.6}$ & 72.1$_{\pm 3.2}$ & 68.4$_{\pm 1.2}$ & 62.7$_{\pm 3.7}$ & 58.5$_{\pm 3.8}$ \\

    \midrule
    A/B prompting & 77.4$_{\pm 3.6}$ & 81.9$^{*}$$_{\pm 3.7}$ & 76.5$_{\pm 4.3}$ & 80.5$_{\pm 3.5}$ & 80.8$^{*}$$_{\pm 1.1}$ & \underline{82.5}$^{*}$$_{\pm 0.3}$ \\
    Yes/No prob comp & 73.6$_{\pm 3.2}$ & $\mathbf{88.9}$$^{*}$$_{\pm 0.3}$ & $\mathbf{84.1}$$^{*}$$_{\pm 1.2}$ & $\mathbf{84.0}$$^{*}$$_{\pm 2.0}$ & $\mathbf{89.0}$$^{*}$$_{\pm 0.2}$ & $\mathbf{86.8}$$^{*}$$_{\pm 0.4}$ \\
    \bottomrule
    \end{tabular}

    \begin{minipage}{\linewidth}
        \centering
        \small
        \vspace{4pt}
        (a) BLiMP
        \vspace{4pt}
    \end{minipage}

    \begin{tabular}{
        >{\hspace{-2pt}}l<{\hspace{-2pt}}
        c<{\hspace{-2pt}}
        c<{\hspace{-2pt}}
        c<{\hspace{-2pt}}
        c<{\hspace{-2pt}}
        c<{\hspace{-2pt}}
        c<{\hspace{-2pt}}
    }
    \toprule
     & Llama-3 & Llama-3-Inst. & Yi-1.5 & Yi-1.5-Chat & Qwen2 & Qwen2-Inst. \\
    \midrule
    LP & \underline{83.2} & 80.4 & \underline{86.8} & \underline{85.3} & 85.4 & \underline{85.4} \\
    MeanLP & 74.5 & 71.7 & 75.9 & 74.5 & 74.5 & 74.3 \\
    PenLP & 80.3 & 77.7 & 84.3 & 82.3 & 82.2 & 82.0 \\

    \midrule
    In-template LP & $\mathbf{85.7}$$^{*}$$_{\pm 0.5}$ & $\mathbf{82.9}$$^{*}$$_{\pm 0.3}$ & $\mathbf{87.4}$$^{*}$$_{\pm 0.4}$ & $\mathbf{86.8}$$^{*}$$_{\pm 0.8}$ & $\mathbf{87.9}$$^{*}$$_{\pm 0.3}$ & $\mathbf{86.2}$$^{*}$$_{\pm 0.3}$ \\
    In-template MeanLP & 79.9$_{\pm 0.9}$ & 77.7$_{\pm 0.6}$ & 78.8$_{\pm 1.3}$ & 79.4$_{\pm 1.2}$ & 77.7$_{\pm 1.2}$ & 77.5$_{\pm 1.4}$ \\
    In-template PenLP & 83.0$_{\pm 0.5}$ & \underline{80.7}$^{*}$$_{\pm 0.4}$ & 84.0$_{\pm 0.8}$ & 83.3$_{\pm 1.1}$ & 83.4$_{\pm 0.4}$ & 82.9$_{\pm 0.5}$ \\
    In-template compar. LP & 68.2$_{\pm 2.4}$ & 58.2$_{\pm 3.0}$ & 63.9$_{\pm 4.6}$ & 61.8$_{\pm 3.4}$ & 68.1$_{\pm 4.5}$ & 60.6$_{\pm 3.9}$ \\

    \midrule
    A/B prompting & 68.0$_{\pm 3.5}$ & 69.2$_{\pm 4.0}$ & 74.2$_{\pm 3.6}$ & 75.5$_{\pm 3.0}$ & 77.3$_{\pm 4.2}$ & 80.9$_{\pm 1.6}$ \\
    Yes/No prob comp & 76.3$_{\pm 1.3}$ & 76.9$_{\pm 0.9}$ & 78.2$_{\pm 0.9}$ & 81.7$_{\pm 0.3}$ & \underline{87.2}$^{*}$$_{\pm 0.4}$ & 83.9$_{\pm 0.4}$ \\
    \bottomrule
    \end{tabular}

    \begin{minipage}{\linewidth}
        \centering
        \small
        \vspace{4pt}
        (b) CLiMP
    \end{minipage}

    \caption{
        Percentage accuracy (averaged over templates) by method and model. $\pm$ denotes standard deviation. The bold font denotes the best scores. Underlines denote the second best. See \cref{sec:max-accuracy} for the max accuracy.
        * denotes scores significantly higher than LP ($p \leq 0.01$).
        % * and ** denote scores significantly higher than LP (p <= 0.05 and p <= 0.01, respectively).
    }
    \label{tab:summary-mean}
\end{table*}

%%%%%%%%%%%%%%%%%%%%

\section{Experimental Setup}

\paragraph{Models}
We use eight state-of-the-art LLMs, among which Llama-3-70B \cite{llama3}, Mixtral-8x7B-v0.1 \cite{jiang2024mixtral}, Qwen2-57B-A14B \cite{qwen1.5}, and Yi-1.5-34B \cite{ai2024yiopenfoundationmodels} are base models, while Llama-3-70B-Instruct, Mixtral-8x7B-Instruct-v0.1, Qwen2-57B-A14B-Instruct, and Yi-1.5-34B-Chat are instruct models based on the pre-trained counterparts.
These models ranked relatively high in the leaderboard of English language understanding\footnote{
    \url{https://paperswithcode.com/sota/multi-task-language-understanding-on-mmlu}
} or Chinese LLMs\footnote{
    \url{https://huggingface.co/spaces/BAAI/open_cn_llm_leaderboard}
} at the time of model selection.
For Chinese experiments, we substituted Yi-1.5 for Mixtral because Mixtral is not explicitly trained for Chinese tasks, while Yi-1.5 ranked high in the Chinese LLM leaderboard.
% これらのモデルは、MMLUで高い精度を達成したmodel familyの、執筆時点での最新版
We hereafter abbreviate these models, e.g., to Llama-3, omitting the model sizes and minor versions.
Post-training for the three instruct models includes supervised fine-tuning on an instruction dataset, i.e., instruction-tuning and DPO.
They are publicly available on Hugging Face Hub.
% Qwen2-57B-A14Bは、CMMLUで、GPT-4や、同等のサイズのモデルを上回る精度88.5を達成している
% https://qwenlm.github.io/blog/qwen2/
% https://github.com/haonan-li/CMMLU/blob/master/README_EN.md
On inference, we perform 4-bit quantization using bitsandbytes\footnote{
    \url{https://github.com/TimDettmers/bitsandbytes}
} to compress the models.
The computational budgets are described in \cref{sec:budgets}.

\paragraph{Benchmarks}
We use two MP acceptability judgment benchmarks: BLiMP \cite{warstadt-etal-2020-blimp} for English and CLiMP \cite{xiang-etal-2021-climp} for Chinese.
BLiMP is composed of minimal pairs from 67 different paradigms, each containing 1,000 pairs of sentences.
The paradigms are grouped into 12 categories of linguistic phenomena.
CLiMP consists of 16 paradigms, each with 1,000 pairs like BLiMP.
The paradigms are grouped into 9 linguistic phenomena.
We focus on these two because no other MP benchmarks contain hundreds or thousands of sentences for each paradigm, to our knowledge, which is important for reliable experiments. % JBLiMPについて、これが真であることは確認済
The linguistic phenomena and licenses of the two benchmarks are detailed in \cref{sec:phenomena} and \cref{sec:urls}, respectively.

\paragraph{Evaluation metric}
We evaluate the methods by accuracy. 
Random chance accuracy is 50\%, as the two classes are balanced in our benchmarks.

%%%%%%%%%%%%%%%%%%%%

\begin{figure*}[!ht]
    \includegraphics[width=1.0\linewidth]{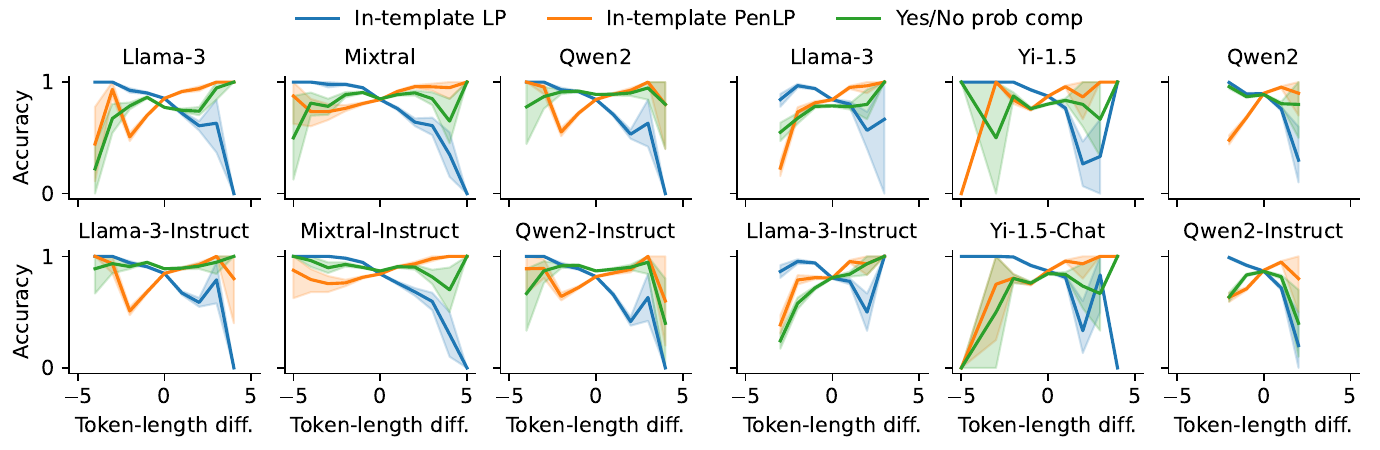}
    
    \begin{minipage}{0.49\linewidth}
        \centering
        \small \qquad (a) BLiMP
    \end{minipage}
    \begin{minipage}{0.49\linewidth}
        \centering
        \small (b) CLiMP
    \end{minipage}
    
    \caption{Top methods' correlation between the token-length difference ($|\bm{s}_{\mathrm{acceptable}}| - |\bm{s}_{\mathrm{unacceptable}}|$) and the accuracy (best template) by model, showing the robustness of Yes/No prob comp. The shadow denotes 95\% confidence intervals.}
    \label{fig:accuracy-by-lendiff}
\end{figure*}

%%%%%%%%%%%%%%%%%%%%

\section{Results}
\label{sec:results}

\cref{tab:summary-mean} summarizes the results.
The statistics of the in-template probability readout methods and prompting methods are the average of the five scores by the five versions of templates.\footnote{
    We conducted paired bootstrap resampling \cite{koehn-2004-statistical} to validate whether each of our methods statistically significantly improves accuracy compared to LP, the best-performing conventional method; we performed 1,000 resamplings with replacement, sampling 67,000 and 16,000 instances each time for BliMP and CLiMP, respectively.
% We performed paired bootstrap resampling to see if our methods significantly improve on LP, i.e., the best-performing conventional method, we performed bootstrap tests.
}
% They show that Yes/No prob comp and in-template LP are particularly strong across languages.
% Both methods are at least the third best for seven out of eight model-benchmark pairs.

Comparison between methods reveals the effectiveness of in-template LP and Yes/No prob comp.
In-template LP achieves significantly higher accuracies than LP in all settings, i.e., benchmark-model pairs, across languages.
On CLiMP, it marks the highest accuracy for all models.
% CLiMPでは、我々の手法はhumanを超えなかったが、CLiMPのhumanは信頼性が若干あやしい問題があるので、この点に踏み込まない
Sentence probability readout methods---LP, MeanLP, and PenLP---underperformed in-template LP in all settings, although they have been dominant in previous studies.
This indicates that including guidance about the task in the input to LLMs improves judgment performance.
Meanwhile, Yes/No prob comp achieves the highest accuracy for five out of six models on BLiMP; the mean accuracies of Llama-3-Instruct and Qwen2 exceed that of humans (the majority vote of 20 crowd workers) reported in \citet{warstadt-etal-2020-blimp}, 88.6\%.

Methods giving two sentences to the model---A/B prompting and in-template comparative LP---consistently underperformed Yes/No prob comp and in-template LP, respectively, suggesting that LLMs struggle to handle choice identifiers such as A and B (See \cref{sec:analyzing-ab-prompting} for more analysis, which reveals the low performance of A/B prompting can be partly attributed to a preference for a specific choice).
% This is notable, given that making LLMs select an identifier from multiple choices is a common prompting approach for classification tasks such as question answering \cite{hendrycks2021measuring}.
This may suggest that making LLMs select an identifier from multiple choices---a common approach for classification tasks such as question answering \cite{hendrycks2021measuring}---is suboptimal for making full use of LLMs' knowledge in general; substituting our choice-free methods could improve performance in other tasks.
% Archive:
% Comparison within the three pairs of the base and instruct models finds that ...
% The performance of in-template LP slightly deteriorates after post-training.
% Yes/No prob comp does not always perform better with an instruct model than with a base model.
% In summary, post-training does not necessarily improve judgment performance.

%%%%%%%%%%%%%%%%%%%%

\section{Analysis}
\label{sec:analysis}

Given the remarkable performance of Yes/No prob comp and in-template LP, this section further investigates their strengths and weaknesses.

%%%%%%%%%%%%%%%%%%%%

\begin{figure*}[ht]
    \centering
    % \begin{minipage}{0.74\linewidth}
    %     \centering
    %     \includegraphics[width=1.0\linewidth]{figures/accuracy_by_phenomenon.pdf}
    % \end{minipage}
    % \begin{minipage}{0.24\linewidth}
    %     \centering
    %     \vspace{11pt}
    %     \includegraphics[width=1.0\linewidth]{figures/accuracy_by_phenomenon_climp.pdf}
    % \end{minipage}
    \includegraphics[width=1.0\linewidth]{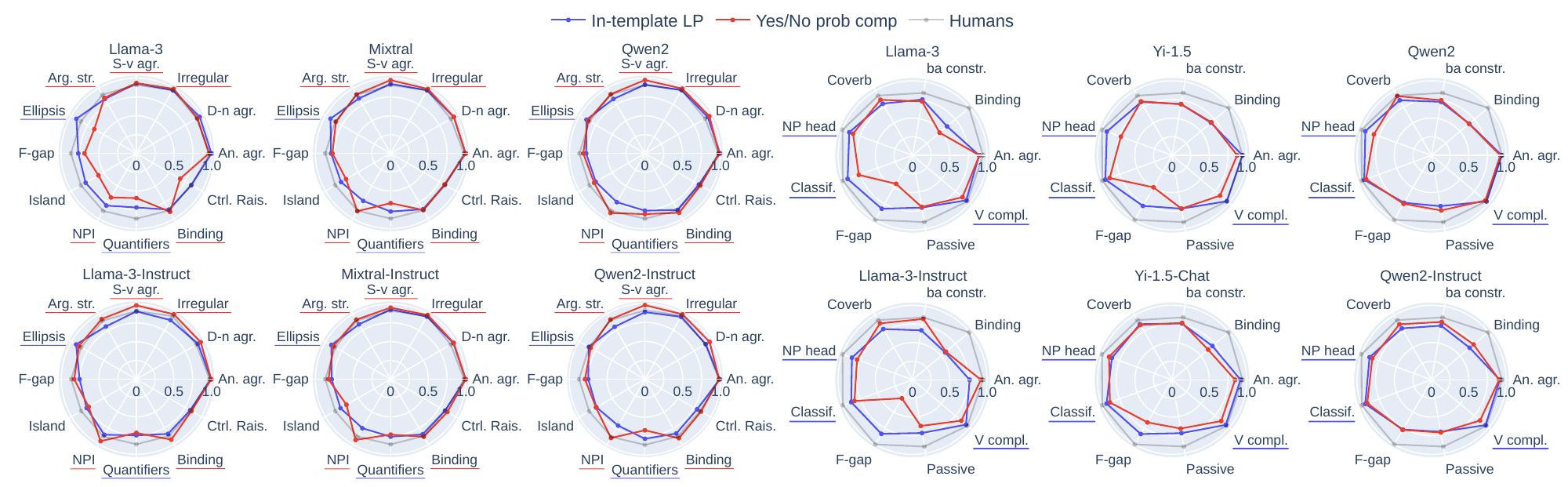} \\
    
    \begin{minipage}{0.49\linewidth}
        \centering
        \small (a) BLiMP
    \end{minipage}
    \begin{minipage}{0.49\linewidth}
        \centering
        \small (b) CLiMP
    \end{minipage}

    \caption{
        Accuracy of top methods (best template) and humans, by linguistic phenomenon and model, demonstrating the difference in strengths between methods. 
        For each benchmark, phenomena where either method wins the other by at least 1 point for at least five models are underlined.
    }
    \label{fig:accuracy-by-phenomenon}
\end{figure*}

%%%%%%%%%%%%%%%%%%%%

\paragraph{Yes/No prob comp is robust against token-length bias.}

\cref{fig:accuracy-by-lendiff} illustrates the correlations between the token-length difference and the accuracy.
The token-length difference is $|\bm{s}_{\mathrm{acceptable}}| - |\bm{s}_{\mathrm{unacceptable}}|$ where $\bm{s}_\mathrm{x}$ denotes the token sequence of either sentence.
A level line denotes that the token-length difference does not affect the method.
Across the models, the following trends are observed. 
(1) The token-length difference biases the readout methods. 
The accuracy of in-template LP decreases as the difference grows because the acceptable sentence is less likely to be given a high probability.
In-template PenLP suffers a reversed tendency; due to normalization, it becomes weaker as the unacceptable sentence gets longer than the acceptable one.
(2) Yes/No prob comp is relatively robust against the bias. 
Its accuracy does not drop as much as that of the other methods, even when the token lengths differ by a large margin.

% https://en.wikipedia.org/wiki/Point-biserial_correlation_coefficient
These observations are quantitatively supported by the correlation coefficient between the token-length difference and the dichotomous variable that gets 1 for a successful prediction and 0 for a failure.
\cref{tab:success-lendiff-correlation} shows the average coefficients of Yes/No prob comp are much closer to zero than those of readout methods on both benchmarks, demonstrating its robustness against the token-length bias.
This, in turn, indicates that the readout methods need better normalization techniques.

\begin{table}[t]
    \centering
    \small
    \begin{tabular}{lcc}
    \toprule
     & BLiMP & CLiMP \\
    \midrule
    % LP & -0.145 \\
    % PenLP & 0.099 \\
    In-template LP & $-0.118$ & $-0.135$ \\
    In-template PenLP & \phantom{$-$}0.094 & \phantom{$-$}0.182 \\
    Yes/No prob comp & $\mathbf{-0.019}$ & \phantom{$-$}$\mathbf{0.051}$ \\
    \bottomrule
    \end{tabular}
    \caption{Top methods' point biserial correlation coefficient between the prediction success and token-length difference (averaged over models) by benchmark. The bold font denotes the value closest to zero.}
    % (the point biserial correlation coefficient is equivalent to the Pearson correlation coefficient)
    \label{tab:success-lendiff-correlation}
\end{table}

%%%%%%%%%%%%%%%%%%%%

\begin{table*}[ht]
    \centering
    \small
    % \begin{tabular}{lcccccccc}
    % \toprule
    %  & \multicolumn{6}{c}{BLiMP} & \multicolumn{2}{c}{CLiMP} \\
    % \cmidrule(rl){2-7} \cmidrule(l){8-9}
    %  & Llama-3 & \makecell{Llama-3\\-Instruct} & Mixtral & \makecell{Mixtral\\-Instruct} & Qwen2 & \makecell{Qwen2\\-Instruct} & Qwen2 & \makecell{Qwen2\\-Instruct} \\

    % \midrule
    % Ensemble P-only & 76.0 & 89.0 & 85.4 & 84.4 & 89.5 & 87.0 & 86.6 & 85.0 \\
    % Ensemble Mix-P3 & 82.7 & $\mathbf{89.7}$ & $\mathbf{87.5}$ & $\mathbf{86.6}$ & $\mathbf{90.2}$ & $\mathbf{87.7}$ & 87.5 & 87.5 \\
    % Ensemble Mix-L3 & $\mathbf{86.1}$ & 86.3 & 86.5 & 86.4 & 86.7 & 84.8 & $\mathbf{89.7}$ & $\mathbf{88.6}$ \\
    % Ensemble L-only & 85.1 & 84.1 & 84.6 & 84.3 & 84.3 & 81.0 & 88.4 & 87.0 \\
    % \midrule
    % In-template LP (oracle) & 85.0 & 84.2 & 84.6 & 84.5 & 84.1 & 81.3 & 88.2 & 86.5 \\
    % Yes/No prob comp (oracle) & 77.8 & 89.3 & 85.6 & \underline{87.5} & 89.2 & 87.4 & 86.6 & 85.2 \\
    % \bottomrule
    % \end{tabular}

    \begin{tabular}{lcccccccc}
    \toprule
     & \multicolumn{4}{c}{BLiMP} & \multicolumn{4}{c}{CLiMP} \\
    \cmidrule(rl){2-5} \cmidrule(l){6-9}
     & Llama-3 & \makecell{Llama-3\\-Instruct} & Qwen2 & \makecell{Qwen2\\-Instruct} & Llama-3 & \makecell{Llama-3\\-Instruct} & Qwen2 & \makecell{Qwen2\\-Instruct} \\

    \midrule
    Ensemble L0:P5 & 76.0 & 89.0 & 89.5 & 87.0 & 76.0 & 76.6 & 87.6 & 84.1 \\
    Ensemble L2:P3 & 82.7 & $\mathbf{89.7}$ & $\mathbf{90.2}$ & $\mathbf{87.7}$ & 79.1 & 78.4 & 89.4 & 87.1 \\
    Ensemble L3:P2 & $\mathbf{86.1}$ & 86.3 & 86.7 & 84.8 & $\mathbf{87.1}$ & $\mathbf{83.8}$ & $\mathbf{90.5}$ & $\mathbf{89.4}$ \\
    Ensemble L5:P0 & 85.1 & 84.1 & 84.3 & 81.0 & 86.1 & 83.0 & 88.4 & 87.0 \\
    \midrule
    In-template LP (oracle) & 85.0 & 84.2 & 84.1 & 81.3 & 86.1 & 83.2 & 88.2 & 86.5 \\
    Yes/No prob comp (oracle) & 77.8 & 89.3 & 89.2 & 87.4 & 78.1 & 78.4 & 87.5 & 84.3 \\
    \bottomrule
    \end{tabular}

    \caption{Percentage accuracy of voting ensembles of in-template LP and Yes/No prob comp, with the oracle (max) accuracy by single methods (best template). The bold font denotes the best ensemble score. Results on Mixtral and Yi models are omitted because the same trend is observed; see \cref{sec:ensemble} for their results.}
    \label{tab:ensemble}
\end{table*}

%%%%%%%%%%%%%%%%%%%%

\begin{figure*}[ht]
    \centering
    \includegraphics[width=1.0\linewidth]{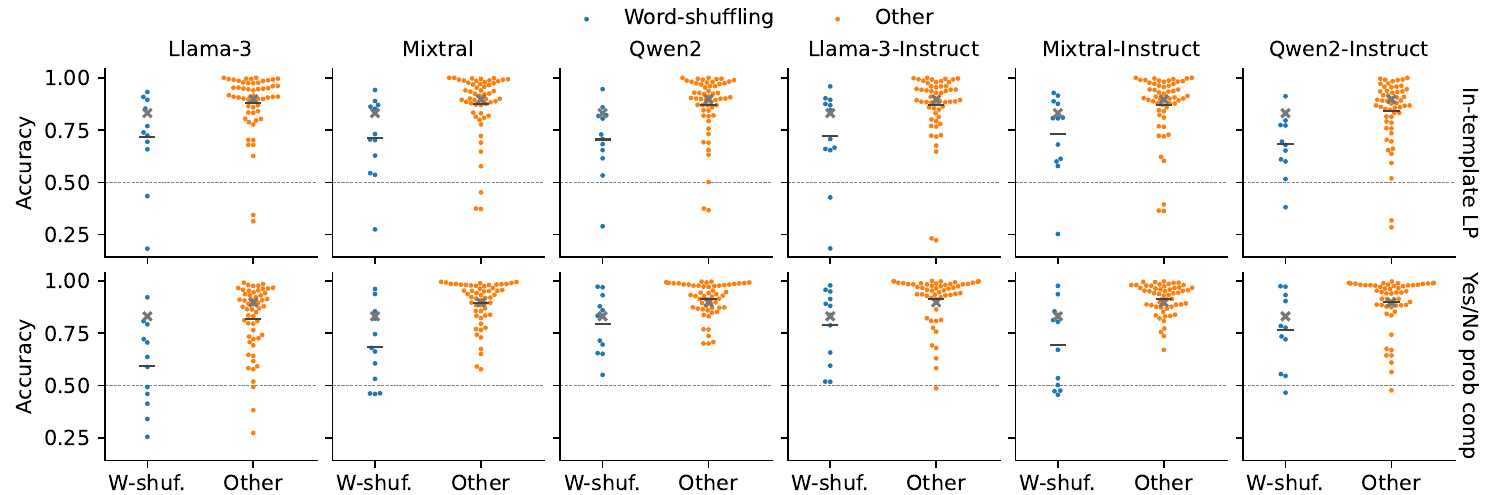}
    \caption{
        Accuracy on BLiMP by paradigm type, method, and model (best template), showing the difficulty of word-shuffling paradigms.
        Each dot represents a paradigm, short bars denote the mean accuracy of the category, cross markers denote mean human accuracies, and dashed lines denote the chance accuracy.
    }
    \label{fig:accuracy-by-bowmatches}
\end{figure*}

%%%%%%%%%%%%%%%%%%%%

\paragraph{In-template readout and Yes/No prob comp excel in different phenomena.}

\cref{fig:accuracy-by-phenomenon} illustrates the accuracy of in-template LP, Yes/No prob comp, and the humans by linguistic phenomenon; the scores of humans are from \citet{warstadt-etal-2020-blimp} and \citet{xiang-etal-2021-climp}.
For in-template LP and Yes/No prob comp, the result of the best-performing template is shown.
Here we find that the two methods have different strengths.
On BLiMP, Yes/No prob comp excels at phenomena such as Subject-verb agreement (S-v agr.) and Binding for most (at least five out of six) models (See \cref{sec:phenomena} for examples of these phenomena).
In contrast, in-template LP is superior in Ellipsis and Quantifiers for most models. 
On CLiMP, Yes/No prob comp is good at Coverb and in-template LP at NP head finality (NP head).
This indicates that each method harnesses different aspects of the models' grammatical knowledge.

Given the aforementioned token-length bias, one hypothesis to explain this difference would be that Yes/No prob comp is more accurate in phenomena with large token-length differences.
Our analysis on BLiMP, however, does not support this, as detailed in \cref{sec:lendiff-accdiff}.

Meanwhile, some phenomena are challenging for both methods.
As \cref{fig:accuracy-by-phenomenon} shows, on BLiMP, the two methods underperform humans for all models in Island effects and Quantifiers, which were shown to be challenging also by \citet{warstadt-etal-2020-blimp}. %, with accuracies of the best model a little above 70\%.
On CLiMP, our methods struggle with phenomena such as Binding and Passive. %, lagging far behind humans.

%%%%%%%%%%%%%%%%%%%%

\paragraph{Voting ensembles of the top two methods further improve the performance.}

Given the different strengths of in-template LP and Yes/No prob comp, we ensemble these methods to see if they can complement each other to achieve higher accuracy.

% Now we have 10 predictions by the two methods, each with five templates.
Now we have 10 sets of predictions by the two methods, as each has five templates.
To compare ensembling single-method predictions and ensembling multi-method predictions on equal terms, we sample five without replacement from the 10 and perform majority voting by the five.
% We prepare the following four settings with different numbers of predictions from each method: L0:P5, L2:P3, L3:P2, L5:P0, where the digits after L and P denote how many predictions by in-template LP and Yes/No prob comp to include, respectively.
% \cref{tab:ensemble} demonstrates that ensembling predictions from the two methods, i.e., L2:P3 or L3:P3, yields the best result for most models, surpassing the max accuracies of methods without ensembling.
% The best accuracy achieved by L2:P3, Qwen2 is over that of humans (described in \cref{sec:results}) by 1.6\%.
We prepare the following four settings, which differ in the balance between the two methods: \textit{P-only}, \textit{Mix-P3}, \textit{Mix-L3}, and \textit{L-only}.
P-only and L-only are ensembles of predictions by Yes/No prob comp only and in-template LP only, respectively.
Mix-P3 and Mix-L3 use three predictions from Yes/No prob comp and in-template LP, respectively, with two predictions from the other method.
We report the mean accuracy of 10 trials for these settings as the result is non-deterministic due to sampling.

\cref{tab:ensemble} demonstrates that ensembles of the two methods, either Mix-P3 or Mix-L3, yield the best results across models, surpassing the oracle (max) accuracies of methods without ensembling, except for Mixtral-Instruct on BLiMP.
The highest score by Mix-P3 with Qwen2 is 1.6 points higher than humans (described in \cref{sec:results}).
This indicates that the two methods have complementary capabilities.
% TODO? camera-ready版で、これが理論的にどのような意味を持つのかについて書く

% 以下の結果は、majority votingの細かい性能は全体の流れの中で見ると重要でないので、一部の結果（islandやquantifiers？）を載せるのみにする
% anaphor_agreement            0.987000
% argument_structure           0.914333
% binding                      0.900714
% control_raising              0.875400
% determiner_noun_agreement    0.989125
% ellipsis                     0.898500
% filler_gap_dependency        0.826286
% irregular_forms              0.994500
% island_effects               0.795500
% npi_licensing                0.935857
% quantifiers                  0.822250
% subject_verb_agreement       0.980000
% overall                      0.902269

%%%%%%%%%%%%%%%%%%%%

\begin{figure*}[t]
    \centering
    \includegraphics[width=1.0\linewidth]{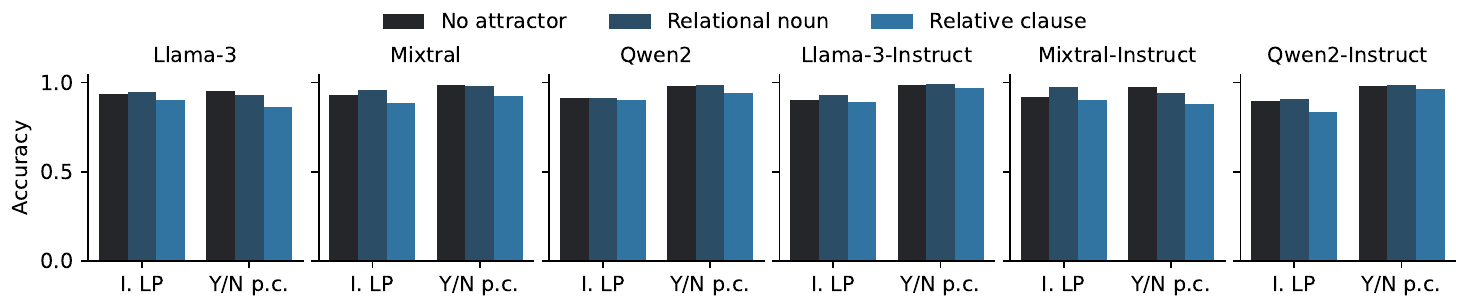}
    \caption{Accuracy on S-v agr. paradigms in BLiMP by attractor type, method, and model (best template), showing that attractors in a relative clause hinder acceptability judgments.}
    \label{fig:accuracy-by-attrtype}
\end{figure*}

%%%%%%%%%%%%%%%%%%%%

\paragraph{Word-shuffling paradigms are challenging.}

BLiMP's 67 paradigms can be divided into two categories based on whether paired sentences of the paradigm have the same bag of words when the cases are ignored.
We call the paradigms where this is true \textit{word-shuffling paradigms}.
In other words, the unacceptable sentence in word-shuffling paradigms can be obtained by shuffling the words in the acceptable counterpart.
Here, we focus on BLiMP because CLiMP does not have word-shuffling paradigms.
Following is an example pair from a word-shuffling paradigm, coordinate\_structure\_constraint\_complex\_left\_branch.
% We want to move the numbers to the left, so use enumerate where enumiitem package is loaded
% The four parameters ending with "sep" remove the vertical spaces
\begin{enumerate}[leftmargin=20pt, topsep=4pt-\parskip, partopsep=0pt, itemsep=0pt, parsep=0pt]
    \item \textit{What reports did Rose hate and James find?}
    \item \textit{*What did Rose hate reports and James find?}
\end{enumerate}

% \begin{asparaenum}[(a)]
%     % \item \textit{Each book is there disturbing Margaret.}\footnote{
%     %     This sentence is non-sensical, which could lower the accuracy of judgments.
%     % }
%     % \item \textit{*\ul{There is each book} disturbing Margaret.}
% \end{asparaenum}

\cref{fig:accuracy-by-bowmatches} shows the accuracy by paradigm, paradigm type—word-shuffling or not, demonstrating that the word-shuffling paradigms have much lower accuracy than other phenomena across methods and models.
The accuracy of word-shuffling paradigms averaged over models and methods is 71.6\% compared to 87.9\% of other paradigms.
The best accuracy on word-shuffling paradigms, achieved by Yes/No prob comp with Qwen2, was 79.6\%, which is much lower than that of the same method on other paradigms, 91.3\%.
% The paradigm marking the lowest accuracy is aforementioned \texttt{existential\_there\_quantifiers\_2}, whose accuracy is only 39.9\% on average.
Note that such large differences are not observed for humans according to the data released by \citet{warstadt-etal-2020-blimp}; humans' accuracy on word-shuffling paradigms and other paradigms are, on average, 83.1\% and 89.7\%, respectively.
This suggests that word-shuffling paradigms remain a challenge for the current LLMs, as they have trouble recognizing word shuffling that corrupts grammar even with our best-performing method.

Why are word-shuffling paradigms difficult?
We hypothesize that the LLMs are insensitive to the word order of the inputs.
Previously, \citet{sinha-etal-2021-unnatural} demonstrated the word-order insensitivity of Transfer-based models, such as RoBERTa \cite{liu2019robertarobustlyoptimizedbert} and BART \cite{lewis-etal-2020-bart}, in the task of natural language inference.
% \citet{sinha-etal-2021-masked} showed permuting word order in the dataset for pre-training has little effect on downstream task performance of RoBERTa \cite{liu2019robertarobustlyoptimizedbert}.
% \citet{papadimitriou-etal-2022-classifying-grammatical} probed GPT-2 to show that the information of syntactic word order is lost when... % They claim slightly different things. They discuss locally-shuffled sentences
\citet{sinha-etal-2021-masked} and \citet{pham-etal-2021-order} report the results to the same effect.
The Transformer architecture, which forms the basis of the models they and we used, may generally struggle to capture the difference in word order.

% As a hypothesis, we conjecture that the LLMs' judgment capabilities hinge on the distributional prior.
% The distributional prior is discussed by \cite{sinha-etal-2021-masked} to explain the result of their experiments.
% permuting word order in the dataset for pre-training has little effect on downstream task performance

%%%%%%%%%%%%%%%%%%%%

\paragraph{Attractors in a relative clause lower the performance.}
% \label{sec:attractors}

% Intervening materials can lower the judgment performance of humans \cite{BOCK199145}.
Attractors refer to material intervening agreement dependencies, and previous work has shown that they can impair acceptability judgments.
Below are examples of different attractor types in S-v agr., from \citet{warstadt-etal-2020-blimp}; (a) contains no attractor, (b) has an attractor as a relational noun, and (c) has an attractor in a relative clause.
Because subject-verb agreement does not exist in Chinese, we focus on English here.
\begin{enumerate}[topsep=4pt-\parskip, partopsep=0pt, itemsep=0pt, parsep=0pt]
    \item \textit{The sisters bake/*bakes.}
    \item \textit{The sisters of Cheryl bake/*bakes.}
    \item \textit{The sisters who met Cheryl bake/*bakes.}
\end{enumerate}
Using such sentence pairs, \citet{warstadt-etal-2020-blimp} and \citet{mueller-etal-2020-cross} investigated the sensitivity of models to mismatches in S-v agr. 
They showed an attractor noun of the opposite number often deteriorates accuracy, particularly when the attractor is a relational clause, as in sentence (c).

\cref{fig:accuracy-by-attrtype} shows both top methods suffer the same issue across models.
The accuracy averaged over methods and models drops from 94.5\% for the agreement with no attractors to 90.4\% for the agreement with attractors in a relative clause.
In contrast, attractors as relational nouns do not necessarily lower the performance.

%%%%%%%%%%%%%%%%%%%%

\section{Conclusion}

We investigate how to derive the most accurate acceptability judgments from LLMs by comparing nine methods, using eight LLMs and two benchmarks.
Our experiments reveal that in-template LP consistently outperforms conventional sentence probability readout methods while Yes/No prob comp achieves the highest accuracies on the English benchmark.
Our analysis demonstrates that the two methods excel in different phenomena, suggesting they harness different aspects of LLMs' grammatical knowledge.
We also find that ensembling the two methods achieves even higher accuracy.
Consequently, we recommend ensembling the two methods or employing in-template LP as more effective alternatives to conventional approaches.
Meanwhile, we show that word-shuffling paradigms remain difficult for all our methods, posing a challenge for future work.

%%%%%%%%%%%%%%%%%%%%

\section{Limitations}

% Acceptable sentences in the two datasets can contain non-sensical ones, as the sentences in BLiMP and CLiMP are automatically generated without their senses being considered.
In BLiMP and CLiMP, acceptable sentences are designed to be grammatical or well-formed, and the sentences in each paradigm were validated by humans \cite{warstadt-etal-2020-blimp,xiang-etal-2021-climp}.
However, acceptable sentences can be non-sensical as they are automatically generated without their senses being considered.
For example, the acceptable sentence in the following pair from the existential\_there\_quantifiers\_2 phenomenon is non-sensical; it is difficult to identify what situation is described in the acceptable sentence, at least without context.
\begin{enumerate}[topsep=4pt-\parskip, partopsep=0pt, itemsep=0pt, parsep=0pt]
    \item \textit{All convertibles weren't there existing.}
    \item \textit{*There weren't all convertibles existing.}
\end{enumerate}
To make correct acceptability judgments, our prompts or in-template inputs guided the LLMs to focus on acceptability or grammatical correctness.
However, this may be insufficient when making judgments on phenomena that contain many non-sensical acceptable sentences like the above.
To address this issue, future work could use prompts or in-template inputs that clearly instruct the model to focus only on the form of each sentence and ignore whether the sentence is sensical.

One of the key findings of this paper is that in-template LP and Yes/No prob comp excel in different linguistic phenomena.
To investigate the reasons for the differences, we examined a hypothesis that Yes/No prob comp is more accurate in phenomena where the acceptable sentence is, on average, longer than the unacceptable one (See \cref{sec:lendiff-accdiff}).
Yet the hypotheses were not supported, leaving the cause of their different strengths an open question.

Throughout the paper, we focused on experiments in the zero-shot setting, aligning the conditions with conventional probability readout methods.
It is notable that some methods nonetheless achieved accuracies surpassing humans.
However, providing few-shot examples in in-template LP and Yes/No prob comp might increase accuracy even further, which is worth investigating in future work.

Additionally, this paper focused on English and Chinese acceptability judgments.
Although this was inevitable to secure a sufficient data size per benchmark, experiments in more languages are desired for more robust conclusions.

%%%%%%%%%%%%%%%%%%%%%%%%%%%%%%%%%%%%%%%%%%%%%%%%%%

\section*{Acknowledgments}

We are grateful to Hongyu Sun for verifying the Chinese translations of the prompts and templates.

\bibliography{anthology,custom}

%%%%%%%%%%%%%%%%%%%%%%%%%%%%%%%%%%%%%%%%%%%%%%%%%%

\newpage
\onecolumn

\appendix

%%%%%%%%%%%%%%%%%%%%

\section{Templates and Prompts}

\begin{table*}[ht]
    \centering
    \small
    \begin{tabular}{l p{0.72\linewidth}}
        \toprule
        Input Type & Example Input \\
        \midrule
        Sentence & \underline{王玉珍震惊了她自己。} \\
        \cmidrule(r){1-1} \cmidrule(l){2-2}
        In-template single & 以下句子在语法上是可以接受的。\textbackslash n\textbackslash n\underline{王玉珍震惊了她自己。} \\
        \cmidrule(r){1-1} \cmidrule(l){2-2}
        \multirow{2}{*}{In-template comparative} & 下面的句子 A 在语法上是可以接受的，而 B 则不可以。\textbackslash n\textbackslash n\newline A：\underline{王玉珍震惊了她自己。}\textbackslash nB：\underline{王玉珍震惊了他自己。} \\
        \bottomrule
    \end{tabular}
    \caption{Example Chinese inputs of the readout methods. The target or inserted sentences are underlined. See \cref{tab:template-settings} for English versions.}
    \label{tab:template-settings-zh}
\end{table*}

\begin{table*}[ht]
    \centering
    \small
    \begin{tabular}{l l p{0.76\linewidth}}
        \toprule
        Type & Role & Example Message \\

        \midrule
        \multirow{3}{*}{A/B} & System & 您的任务是评估给定文本的质量。 \\
        \cmidrule(rl){2-2} \cmidrule(l){3-3}
        & \multirow{2}{*}{User} & 下列句子中，一个符合语法要求，另一个不符合语法要求。哪个可以接受？请用 A 或 B 作答。\textbackslash n\textbackslash nA：\underline{王玉珍震惊了她自己。}\textbackslash nB：\underline{王玉珍震惊了他自己。} \\

        \midrule
        \multirow{2}{*}{Yes/No} & System & 您的任务是评估给定文本的质量。 \\
        \cmidrule(rl){2-2} \cmidrule(l){3-3}
        & User & 下面的句子符合语法要求吗？请用“是”或“否”回答。\textbackslash n\textbackslash n\underline{王玉珍震惊了她自己。} \\
        \bottomrule
    \end{tabular}
    \caption{Example Chinese messages for prompting. The target or inserted sentences are underlined. See \cref{tab:prompts} for English versions.}
    \label{tab:prompts-zh}
\end{table*}

%%%%%%%%%%%%%%%%%%%%

\section{Benchmarks}

\subsection{Linguistic Phenomena}
\label{sec:phenomena}

\begin{table*}[ht]
    \small
    \centering
    \begin{tabular}{llp{0.3\linewidth}p{0.3\linewidth}}
        \toprule
        Field & Phenomenon & Acceptable Example & Unacceptable Example\\
        
        \midrule
        \multirow{4}{*}{Morphology} & Anaphor agr. & \textit{Many girls insulted \underline{themselves}.} & \textit{Many girls insulted \underline{herself}.}\\
         & Det.-noun agr. & \textit{Rachelle had bought that \underline{chair}.} & \textit{Rachelle had bought that \underline{chairs}.}\\
         & Irregular forms & \textit{Aaron \underline{broke} the unicycle.} & \textit{Aaron \underline{broken} the unicycle.}\\
         & Subject-verb agr. & \textit{These casseroles \underline{disgust} Kayla.} & \textit{These casseroles \underline{disgusts} Kayla.}\\
        
        % \cmidrule(r){1-1} \cmidrule(l){2-4}
        \midrule
        \multirow{5}{*}{Syntax} & Arg. structure & \textit{Rose wasn't \underline{disturbing} Mark.} & \textit{Rose wasn't \underline{boasting} Mark.}\\
         & \multirow{2}{*}{Ellipsis} & \textit{Anne's doctor cleans one \underline{important} book and Stacey cleans a few.} & \textit{Anne's doctor cleans one book and Stacey cleans a few \underline{important}.}\\
         & Filler-gap & \textit{Brett knew \underline{what} many waiters find.} & \textit{Brett knew \underline{that} many waiters find.}\\
         & Island effects & \textit{Which \underline{bikes} is John fixing?} & \textit{Which is John fixing \underline{bikes}?}\\

        \midrule
        \multirow{2}{*}{Semantics} & NPI licensing & \textit{The truck has \underline{clearly} tipped over.} & \textit{The truck has \underline{ever} tipped over.}\\
         & Quantifiers & \textit{No boy knew \underline{fewer than} six guys.} & \textit{No boy knew \underline{at most} six guys.}\\
        
        \midrule
        \multirow{2}{*}{Syn. \& Sem.} & Binding & \textit{Carlos said that Lori helped \underline{him}.} & \textit{Carlos said that Lori helped  \underline{himself}.} \\
         & Control/raising & \textit{There was \underline{bound} to be a fish escaping.} & \textit{There was \underline{unable} to be a fish escaping.}\\
        \bottomrule
    \end{tabular}
    \caption{Minimal pairs from each of the twelve linguistic phenomena covered by BLiMP. Differences are underlined.}
    \label{tab:phenomena}
\end{table*}

\begin{table*}[!ht]
    \centering
    \newcommand{\tabincell}[2]{\begin{tabular}  
    {@{}#1@{}}#2\end{tabular}} 
    \small
    
    \resizebox{\textwidth}{!}{% 
    \begin{tabular}{llll}
    \toprule
    Phenomenon & Acceptable Example & Unacceptable Example\\
    \midrule
    
    \tabincell{l}{Anaphor\\agreement} 
    & \tabincell{l}{\small{王玉珍 \ \ 震惊-了\quad \underline{她自己}。}\\
    \small{Jane.F \quad shock-PST \ \underline{herself}.}\\
    \emph{\small{'Jane shocked \underline{herself}.'}}} &\tabincell{l}{\small{王玉珍 \  \ 震惊-了\quad \underline{他自己}。}\\
    \small{Jane.F \quad shock-PST \ \underline{himself}.}\\
    \emph{\small{'Jane shocked \underline{himself}.'}}}\\
    
    Binding 
    & \tabincell{l}{\small{杨颖 \ \ 治疗 吴宇涛 之后 \ 佩服-过 \quad \underline{她自己}。}\\
    \small{Yang.F\ cure \ \  Wu.M\ \  after \ admire-PST \underline{herself}}\\
    \emph{\small{'Yang admired \underline{herself} after she cured Wu.'}}} 
    & \tabincell{l}{\small{杨颖 \quad 治疗 吴宇涛 之后 \  \ 佩服-过 \qquad \underline{他自己}。}\\
    \small{Yang.F \ \ cure \ \  Wu.M\ \   after \ admire-PST \ \ \underline{himself}}\\
    \emph{\small{'Yang admired \underline{himself} after she cured Wu.'}}}  \\
    
    \tabincell{l}{\emph{b\v{a}}\\construction} & \tabincell{l}{\small{王鑫 \qquad \underline{把} \ 自行车 \ 扔  \quad 了。}\\
    \small{Wong.M \ \underline{BA} \ \ bike \ throw \ PST}\\
    \emph{\small{'Wong threw away the bike.'}}} &
    \tabincell{l}{\small{王鑫 \qquad \underline{被} \ \ 自行车 \ 扔 \quad  了 。}\\
    \small{Wong.M \ \underline{PASS}  \ bike \ throw \ PST }\\
    \emph{\small{'Wong \underline{was thrown away} by the bike.'}}}\\
    
    Coverb  & \tabincell{l}{\small{李文清 \quad \underline{乘} \ 卡车 \quad到达-了\quad 咖啡店。}\\
    \small{Lee.M \quad \ \underline{ride} truck \ arrive-PST coffee shop}\\
    \emph{\small{'Lee went to the coffee shop \underline{by} truck.'}}} &
    \tabincell{l}{\small{李文清 \quad \underline{于} 卡车 \quad到达-了\quad 咖啡店。}\\
    \small{Lee.M \quad \underline{at}\quad truck \ \ arrive-PST coffee shop}\\
    \emph{\small{'Lee went to the coffee shop \underline{at} truck.'}}}\\
    
    NP head finality & \tabincell{l}{\small{王梦\qquad 正在 \ 卖\ \underline{张红梅 \ 清洗-过-的}\quad 推车。}\\
    \small{Wong.F PROG sell \underline{May.F clean-PRF-ADJ} trolley}\\
    \emph{\small{‘Wong is selling the trolley \underline{that Mel has cleaned}.’}}} &
    \tabincell{l}{\small{王梦\qquad 正在 \ 卖\ \ 推车 \quad \underline{张红梅 \ 清洗-过-的}。}\\
    \small{Wong.F PROG sell trolley \underline{May.F clean-PRF-ADJ}}\\
    \emph{\small{‘Wong is selling the trolley \underline{that Mel has cleaned}.’}}} \\
    
    Classifier & \tabincell{l}{\small{张杰\quad 正在\ \ 穿过\ \ 一\qquad \qquad \underline{家}\qquad \qquad 艺术画廊}。\\
    
    \small{Jay.M PROG pass one \underline{CL:INSTITUTION} art gallery}\\
    \emph{\small{'Jay is passing through \underline{an} art gallery.'}}} & 
    \tabincell{l}{\small{张杰\ 正在\ \ 穿过\ \ 一\qquad  \quad \underline{段}\qquad \quad 艺术画廊}。\\
    \small{Jay.M PROG pass one \underline{CL:LENGTH} art gallery}\\
    \emph{\small{'Jay is passing through \underline{an} art gallery.'}}}  \\
    
    Filler gap & \tabincell{l}{\small{图书馆，\quad 我 \ 开车 去-过 \ \underline{这个地方}。}\\
    \small{The library, \ \  I \ drive \  to-PRF \underline{this place}}\\
    \emph{\small{‘The library, I have driven to \underline{this place}.’}}} & 
    \tabincell{l}{\small{图书馆，\quad \ \ 我 \ 开车 去-过 \quad  \underline{博物馆}。}\\
    \small{The library, \quad I \ drive \ to-PRF \underline{the museum}}\\
    \emph{\small{‘The library, I have driven to \underline{ the museum}.’}}}  \\
    
    Passive &\tabincell{l}{\small{这些\quad 患者 \quad 被\qquad \underline{转移}-了。}\\
    \small{These patient PASS \underline{transfer}-PST}\\
    \emph{\small{'These patients were \underline{transferred}.'}}}&
    \tabincell{l}{\small{这些\quad 患者 \quad 被\quad \underline{下降}-了。}\\
    \small{These patient PASS \ \underline{fall}-PST}\\
    \emph{\small{'These patients were \underline{fell}.'}}}\\
    
    \tabincell{l}{Verb\\complement} & \tabincell{l}{
    \small{王慧 \qquad 的\quad 文章 \quad 吓 \qquad \underline{坏} \quad 了 \ \  包曼玉 。}\\
    \small{Wong.F POSS article frighten \underline{badly} PST \ \ Bao.F.}\\
    \emph{\small{'Wong's article frightened Bao \underline{badly}.'}}} & 
    \tabincell{l}{
    \small{王慧 \qquad 的\quad 文章 \quad 吓 \qquad \underline{开} \quad 了 \ \  包曼玉 。}\\
    \small{Wong.F POSS article frighten \underline{openly} PST \ \ Bao.F.}\\
    \emph{\small{'Wong's article frightened Bao \underline{openly}.'}}} \\
    
    \bottomrule
    \end{tabular}
    }
    \caption{Minimal pairs from each of the nine linguistic phenomena covered by CLiMP. Differences are underlined. The second line of each example shows a gloss, and the third line is an English translation.}
    \label{tab:phenomena-climp}
\end{table*}

\newpage
\subsection{URLs and Licenses}
\label{sec:urls}

\begin{table}[ht]
    \centering
    \small
    \begin{tabular}{
        p{0.08\linewidth}
        p{0.18\linewidth}
        p{0.40\linewidth}
        p{0.14\linewidth}
        % >{\hspace{-3pt}}
        % >{\raggedright\arraybackslash}p{0.07\linewidth}<{\hspace{-3pt}}
        % >{\raggedright\arraybackslash}p{0.12\linewidth}<{\hspace{-3pt}}
        % >{\raggedright\arraybackslash}p{0.24\linewidth}<{\hspace{-3pt}}
        % >{\raggedright\arraybackslash}p{0.15\linewidth}<{\hspace{-3pt}}
        % >{\raggedright\arraybackslash}p{0.28\linewidth}<{\hspace{-3pt}}
    }
        \toprule
        Name & Paper & URL & License \\
        \midrule
        BLiMP & \citet{warstadt-etal-2020-blimp} & \url{https://github.com/alexwarstadt/blimp} & CC-BY \\
        % \cmidrule(r){1-1} \cmidrule(l){2-4}
        CLiMP & \citet{xiang-etal-2021-climp} & \url{https://github.com/beileixiang/CLiMP} & Not articulated \\
        \bottomrule
    \end{tabular}
    \caption{URLs and licenses of the benchmarks.}
    \label{tab:data_sources}
\end{table}

%%%%%%%%%%%%%%%%%%%%

\section{Computational Budgets}
\label{sec:budgets}

For each experiment on a method or a combination of a method and template, we used a single NVIDIA A6000 GPU with 48GB RAM.
The total GPU hours are estimated to be about 126 hours and 21 hours for the BLiMP and CLiMP experiments, respectively.
\vspace{1em}

%%%%%%%%%%%%%%%%%%%%

\newpage
\section{Results and Analysis}

\subsection{Max Accuracy}
\label{sec:max-accuracy}

\begin{table*}[ht]
    \centering
    \small
    \begin{tabular}{lcccccccc}
    \toprule
     & Llama-3 & Llama-3-Instruct & Mixtral & Mixtral-Instruct & Qwen2 & Qwen2-Instruct \\

    \midrule
    LP & 79.6 & 77.1 & 82.5 & 82.3 & 80.4 & 79.7 \\
    MeanLP & 77.1 & 74.8 & 79.6 & 79.4 & 77.7 & 77.1 \\
    PenLP & 79.2 & 76.8 & 82.2 & 82.0 & 79.9 & 79.2 \\

    \midrule
    In-template LP & $\mathbf{85.0}$ & \underline{84.2} & \underline{84.6} & \underline{84.5} & \underline{84.1} & 81.3 \\
    In-template MeanLP & 83.1 & 82.6 & 83.2 & 83.1 & 82.8 & 80.5 \\
    In-template PenLP & \underline{84.4} & 83.5 & 84.5 & 84.3 & 83.6 & 81.2 \\
    In-template compar. LP & 76.4 & 66.0 & 75.2 & 69.8 & 67.7 & 63.6 \\

    \midrule
    A/B prompting & 79.5 & 83.9 & 81.9 & 83.6 & 81.5 & \underline{82.8} \\
    Yes/No prob comp & 77.8 & $\mathbf{89.3}$ & $\mathbf{85.6}$ & $\mathbf{87.5}$ & $\mathbf{89.2}$ & $\mathbf{87.4}$ \\
    \bottomrule
    \end{tabular}

    \begin{minipage}{\linewidth}
        \centering
        \small
        \vspace{4pt}
        (a) BLiMP
        \vspace{4pt}
    \end{minipage}

    \begin{tabular}{lcccccccc}
    \toprule
     & Llama-3 & Llama-3-Instruct & Yi-1.5 & Yi-1.5-Chat & Qwen2 & Qwen2-Instruct \\

    \midrule
    LP & 83.2 & 80.4 & \underline{86.8} & \underline{85.3} & 85.4 & \underline{85.4} \\
    MeanLP & 74.5 & 71.7 & 75.9 & 74.5 & 74.5 & 74.3 \\
    PenLP & 80.3 & 77.7 & 84.3 & 82.3 & 82.2 & 82.0 \\

    \midrule
    In-template LP & $\mathbf{86.1}$ & $\mathbf{83.2}$ & $\mathbf{87.9}$ & $\mathbf{87.8}$ & $\mathbf{88.2}$ & $\mathbf{86.5}$ \\
    In-template MeanLP & 80.7 & 78.5 & 80.4 & 81.0 & 78.9 & 79.0 \\
    In-template PenLP & \underline{83.6} & \underline{81.1} & 85.1 & 84.7 & 83.8 & 83.5 \\
    In-template compar. LP & 72.2 & 61.6 & 68.2 & 67.3 & 74.2 & 63.5 \\

    \midrule
    A/B prompting & 71.7 & 74.2 & 78.6 & 79.7 & 83.2 & 82.5 \\
    Yes/No prob comp & 78.1 & 78.4 & 79.7 & 82.1 & \underline{87.5} & 84.3 \\
    \bottomrule
    \end{tabular}

    \begin{minipage}{\linewidth}
        \centering
        \small
        \vspace{4pt}
        (a) CLiMP
        \vspace{4pt}
    \end{minipage}

    \caption{
        Percentage max accuracy by method and model.
        The bold font denotes the best scores. Underlines denote the second best.
    }
    \label{tab:summary-max}
\end{table*}

%%%%%%%%%%%%%%%%%%%%

\subsection{Why A/B prompting does not perform well}
\label{sec:analyzing-ab-prompting}

\begin{table}[ht]
    \centering
    \small

    \begin{tabular}{
        c<{\hspace{-2pt}}
        c<{\hspace{-2pt}}
        c<{\hspace{-2pt}}
        c<{\hspace{-2pt}}
        c<{\hspace{-2pt}}
        c<{\hspace{-2pt}}
        c<{\hspace{-2pt}}
        c<{\hspace{-2pt}}
        c<{\hspace{-2pt}}
        c<{\hspace{-2pt}}
        c<{\hspace{-2pt}}
        c<{\hspace{-2pt}}
        c<{\hspace{-2pt}}
        c<{\hspace{-2pt}}
        c<{\hspace{-2pt}}
        c
    }

    \toprule
    \multicolumn{6}{c}{BLiMP} & \multicolumn{6}{c}{CLiMP} \\
    \cmidrule(r){1-6} \cmidrule(l){7-12}
    Llama-3 & \makecell{Llama-3\\-Inst.} & Mixtral & \makecell{Mixtral\\-Inst.} & Qwen2 & \makecell{Qwen2\\-Inst.} & Llama-3 & \makecell{Llama-3\\-Inst.} & Yi-1.5 & \makecell{Yi-1.5\\-Chat} & Qwen2 & \makecell{Qwen2\\-Inst.} \\

    \midrule
    55.0 & 56.6 & 70.1 & 45.9 & 54.0 & 45.7 & 70.5 & 72.8 & 52.7 & 46.7 & 35.9 & 46.1 \\
    \bottomrule
    \end{tabular}

    \caption{Percentage proportion of A in the predictions of A/B prompting (averaged over templates) by model.}
    \label{tab:a-proportion}
\end{table}

The low performance of A/B prompting can be partly attributed to a preference for a specific choice identifier, A or B.
\cref{tab:a-proportion} shows all models except for Yi-1.5 models are at least 7 points more likely to predict one choice over the other one, even though the gold label is sampled from a uniform distribution. % the gold labels are uniformly distributed
This suggests that the current LLMs suffer from selection bias on multiple choices as argued by \citet{zheng2024large}.

%%%%%%%%%%%%%%%%%%%%

\newpage

\subsection{Is Yes/No prob comp strong where the acceptable sentence is longer than the unacceptable?}
\label{sec:lendiff-accdiff}

\begin{figure*}[ht]
    \centering
    \includegraphics[width=1.0\linewidth]{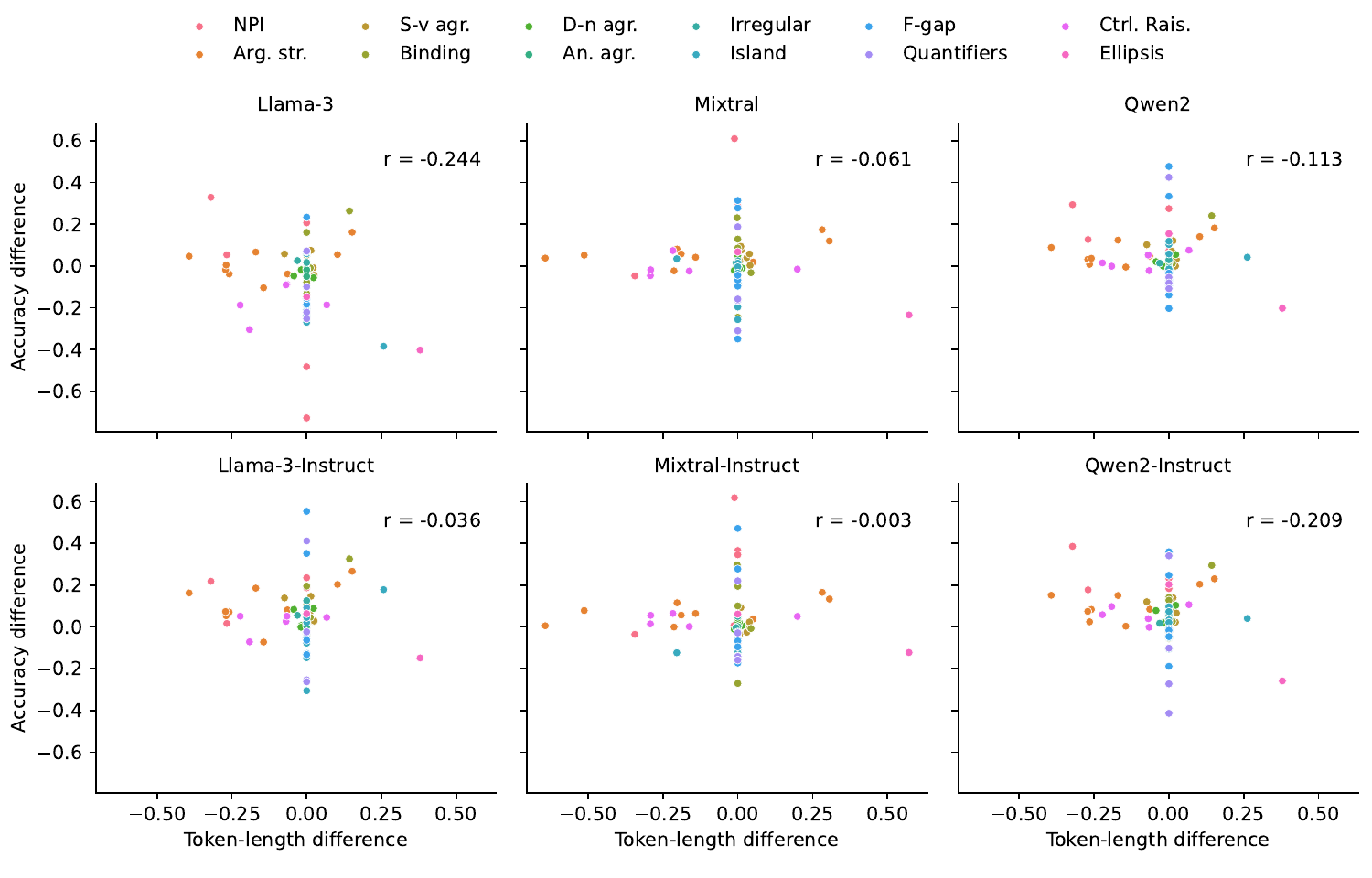}
    \caption{Correlation between the token-length difference ($|\bm{s}_{\mathrm{acceptable}}| - |\bm{s}_{\mathrm{unacceptable}}|$) and the accuracy difference ($\mathrm{accuracy}_\mathrm{\ Yes/No\ prob\ comp} - \mathrm{accuracy}_\mathrm{\ In-template\ LP}$) by model. Each dot represents a paradigm. Plots are annotated with the Pearson correlation coefficient $r$.}
    \label{fig:lendiff-accdiff-corr}
\end{figure*}

\cref{fig:lendiff-accdiff-corr} shows that Yes/No prob comp is not particularly accurate compared to in-template LP in phenomena where the acceptable sentence is, on average, longer than the unacceptable one.
We only find no or weak negative correlations between the accuracy difference and token-length difference.

%%%%%%%%%%%%%%%%%%%%

\newpage

\subsection{Voting Ensembles}
\label{sec:ensemble}

\begin{table}[ht]
    \centering
    \small
    \begin{tabular}{lcccccc}
    \toprule
     & Llama-3 & Llama-3-Instruct & Mixtral & Mixtral-Instruct & Qwen2 & Qwen2-Instruct \\
    \midrule
    Ensemble P-only & 76.0 & 89.0 & 85.4 & 84.4 & 89.5 & 87.0 \\
    Ensemble Mix-P3 & 82.7 & $\mathbf{89.7}$ & $\mathbf{87.5}$ & $\mathbf{86.6}$ & $\mathbf{90.2}$ & $\mathbf{87.7}$ \\
    Ensemble Mix-L3 & $\mathbf{86.1}$ & 86.3 & 86.5 & 86.4 & 86.7 & 84.8 \\
    Ensemble L-only & 85.1 & 84.1 & 84.6 & 84.3 & 84.3 & 81.0 \\
    \midrule
    In-template LP (oracle) & 85.0 & 84.2 & 84.6 & 84.5 & 84.1 & 81.3 \\
    Yes/No prob comp (oracle) & 77.8 & 89.3 & 85.6 & \underline{87.5} & 89.2 & 87.4 \\
    \bottomrule
    \end{tabular}

    \begin{minipage}{\linewidth}
        \centering
        \small
        \vspace{4pt}
        (a) BLiMP
        \vspace{4pt}
    \end{minipage}

    \begin{tabular}{lcccccc}
    \toprule
     & Llama-3 & Llama-3-Instruct & Yi-1.5 & Yi-1.5-Chat & Qwen2 & Qwen2-Instruct \\
    \midrule
    Ensemble P-only & 76.0 & 76.6 & 78.4 & 81.9 & 87.6 & 84.1 \\
    Ensemble Mix-P3 & 79.1 & 78.4 & 79.9 & 83.9 & 89.4 & 87.1 \\
    Ensemble Mix-L3 & $\mathbf{87.1}$ & $\mathbf{83.8}$ & $\mathbf{88.7}$ & $\mathbf{88.9}$ & $\mathbf{90.5}$ & $\mathbf{89.4}$ \\
    Ensemble L-only & 86.1 & 83.0 & 88.0 & 87.7 & 88.4 & 87.0 \\
    \midrule
    In-template LP (oracle) & 86.1 & 83.2 & 87.9 & 87.8 & 88.2 & 86.5 \\
    Yes/No prob comp (oracle) & 78.1 & 78.4 & 79.7 & 82.1 & 87.5 & 84.3 \\
    \bottomrule
    \end{tabular}

    \begin{minipage}{\linewidth}
        \centering
        \small
        \vspace{4pt}
        (b) CLiMP
        \vspace{4pt}
    \end{minipage}
    \caption{Percentage accuracy of voting ensembles of in-template LP and Yes/No prob comp, with the oracle (max) accuracy by single methods (best template). The bold font denotes the best ensemble score. Underlines denote oracle results surpassing the best ensemble result.}
    \label{tab:ensemble-whole}
\end{table}

%%%%%%%%%%%%%%%%%%%%

\end{CJK}
\end{document}